\documentclass[runningheads]{llncs}

\usepackage{eccv}

\usepackage{eccvabbrv}

\usepackage{graphicx}
\usepackage{booktabs}
\usepackage{graphicx}

\usepackage[accsupp]{axessibility}  

\usepackage{hyperref}

\usepackage{orcidlink}

\begin{document}

\title{YOLO-PVC: 2D-to-3D Consolidation of Slice-wise Detections for Volumetric Liver Tumor Localization in MRI} 

\titlerunning{YOLO-PVC}

\author{Talha Waqas\inst{1,2} \and
Mounir Lahlouh\inst{1,3} \and
Kawther Taibouni\inst{1} \and
Mahnoor Waqas\inst{4} \and
Salar Ahmed\inst{5} \and
S\'ebastien Mul\'e\inst{6} \and
Yasmina Leroul-Chenoune\inst{1,2}}

\authorrunning{T.~Waqas et al.}

\institute{
ESME Research Lab, Paris, France \and
LISSI (EA 3956), Universit\'e Paris-Est Cr\'eteil, Vitry-sur-Seine, France \and
LRE EPITA, Le Kremlin-Bic\^etre, France \and
Institute of Space Technology, Islamabad, Pakistan \and
AgileLoop.AI, Islamabad, Pakistan \and
Henri Mondor University Hospital, INSERM IMRB, U955, Cr\'eteil, France
}

\maketitle

\begin{abstract}
  Slice-wise 2D object detectors are increasingly applied to volumetric data due to their computational efficiency and scalability, yet they often yield fragmented and unstable predictions along the depth axis. We propose YOLO-PVC, a lightweight and model-agnostic framework for 2D-to-3D consolidation of slice-wise detections. The method enforces depth continuity, aggregates bounding box coordinates using robust percentile statistics, and further refines axial extent through a lightweight MLP-based calibration module. Unlike naïve stacking or averaging strategies, YOLO-PVC explicitly addresses missing detections and outlier slices along the depth dimension. Experiments on 3D liver MRI volumes across three tumor categories demonstrate consistent improvements over multiple aggregation baselines. The heuristic PVC achieves an overall $\mathrm{IoU}_{3D}$ of $0.665$, while the calibrated variant further improves performance to $0.710$, with high planar overlap ($\mathrm{BEV\ IoU} \approx 0.78$). These results demonstrate that structured geometric consolidation provides an effective and practical solution for volumetric liver tumor localization in clinical MRI.

\end{abstract}

\section{Introduction}
\label{sec:intro}

Volumetric object detection is a fundamental problem in computer
vision with critical applications in medical imaging, where accurate
three-dimensional localization is essential for diagnosis, treatment
planning, and longitudinal monitoring~\cite{Kern2020}. Reliable 3D bounding boxes enable quantitative tumor measurement, inter-scan comparison, and therapy response assessment, making robust volumetric localization clinically meaningful.

Primary liver cancers are among the leading causes of cancer-related
mortality worldwide~\cite{Sung2021}. Hepatocellular carcinoma (HCC) and intrahepatic cholangiocarcinoma (CCA) are the two dominant types, while combined hepatocellular--cholangiocarcinoma (cHCC-CCA), also known as the Mixed tumor, represents a rarer but clinically important
entity~\cite{Forner2018}\cite{Banales2020}\cite{Brunt2018}. These tumors exhibit heterogeneous morphology and variable spatial extent across multi-phase MRI scans, making accurate cross-slice volumetric
localization essential for downstream analysis and longitudinal
evaluation~\cite{Liu2024RadiomicsHCC}.

Most volumetric detection pipelines rely on fully 3D convolutional
networks~\cite{Cicek2016}. While effective in modeling inter-slice
context, such models require dense 3D annotations, large datasets,
and substantial GPU memory, limiting scalability and deployment in
resource-constrained clinical settings~\cite{Zhang2022}.

In contrast, modern 2D object detectors such as YOLO are
computationally efficient and robust~\cite{Redmon2016}. Applied
slice-wise, they produce independent 2D detections along the depth
axis. However, naive stacking of slice-level predictions leads to
fragmented 3D boxes, depth inconsistency, and sensitivity to false
positives and missing slices~\cite{Zhang2022}, resulting in
anatomically implausible volumetric estimates.

This raises a central question: how can reliable and anatomically
consistent 3D bounding boxes be recovered from noisy slice-wise
detections without resorting to heavy 3D architectures?

To address this, we propose YOLO-PVC, where PVC denotes
\emph{Percentile-based Volumetric Consolidation}, a lightweight and
model-agnostic framework for robust 2D-to-3D aggregation. The method
enforces depth continuity constraints and applies robust percentile
statistics to fuse spatial coordinates across slices, explicitly
modeling outliers and missing detections. A shallow MLP-based
calibration module further refines axial extent using slice-level
geometric and confidence features while preserving modularity.
Importantly, YOLO-PVC operates on 3D bounding boxes derived from
expert radiologist segmentation delineations, without requiring
voxel-level supervision at training or inference time. This design
targets the practical clinical setting where only lightweight 2D
detectors and bounding-box labels are available, a common constraint
in resource-limited environments due to high annotation costs and GPU
memory requirements.

Unlike heuristic stacking or simple averaging strategies, YOLO-PVC
provides a structured consolidation mechanism bridging efficient 2D
detection and reliable 3D localization. In this study, we focus on
improving 2D-to-3D stacking pipelines rather than competing with
full 3D detection networks, and our comparisons are made against
alternative 2D stacking and aggregation techniques. Experiments on
arterial-phase liver MRI volumes spanning three tumor categories
demonstrate that robust aggregation substantially improves volumetric
IoU$_{3D}$ while maintaining computational efficiency.

\section{Related Work}
\label{sec:relatedwork}
Volumetric object detection in medical imaging is traditionally
addressed using fully 3D convolutional architectures such as 3D U-Net
variants, V-Net~\cite{Cicek2016} ~\cite{Milletar2016VNetFC}, and volumetric encoder--decoder detection frameworks~\cite{Wittmann2022FocusedDE}. These models capture inter-slice context directly through 3D convolutions and are widely used for segmentation-driven or box-based localization. While effective in modeling spatial continuity, 3D approaches require dense volumetric annotations and substantial GPU memory~\cite{Wang2022}. Their computational cost increases rapidly with spatial resolution and depth, which limits scalability for high-resolution multi-phase MRI and real-time clinical deployment.

To reduce computational burden, an alternative paradigm applies 2D
object detectors slice-wise to volumetric data. Image-based detectors
such as YOLO can be trained efficiently and deployed independently on
each slice~\cite{Redmon2016}, enabling scalable processing of large
volumes. However, slice-wise inference produces predictions that are
independent along the depth axis. Without explicit cross-slice
modeling, volumetric reconstructions often suffer from fragmentation,
inconsistent axial extent, and instability near object
boundaries~\cite{Kumar2024}.

Given slice-level detections, a consolidation step is required to
construct a coherent 3D bounding box~\cite{Kern2020}. In practice,
several geometric stacking strategies are commonly used, which we
summarize below.

\emph{Confidence-weighted averaging} aggregates slice-wise box
coordinates using detection confidence as
weights~\cite{Solovyev2019WeightedBF}. However, this approach does
not explicitly enforce inter-slice continuity and may drift when
detections fluctuate along depth. \emph{Trimmed mean aggregation}
removes extreme coordinate values before averaging in order to
suppress outlier slices~\cite{Wu2022SliceFusion}. While more robust
than simple averaging, it still treats depth extent as a passive
byproduct of detections rather than a structured estimation problem.
\emph{Median fusion} computes coordinate-wise medians across
slices~\cite{Kern2020}. This is robust to extreme values but can
underestimate object extent when boundary slices are inconsistently
detected. \emph{Continuity-aware linking} first associates detections
across adjacent slices using overlap or proximity constraints, forming
a contiguous run of detections before fusion~\cite{deVos2017ConvNetBasedLO}.
This explicitly enforces inter-slice consistency, yet it does not
inherently correct systematic depth bias. \emph{Min--Max stacking}
defines the final 3D box by taking coordinate extrema across the
linked run, effectively enclosing all detected slices~\cite{Wang2023ObjectAsQuery}.
This strategy often preserves lateral coverage but remains sensitive
to boundary noise and spurious detections.

These five stacking strategies represent common geometric heuristics
for slice-to-volume consolidation and serve as baselines in our
experimental evaluation. Although straightforward, they treat
consolidation as a fixed post-processing rule rather than a structured
estimation task and lack mechanisms for adaptive axial refinement.

In contrast to fixed geometric stacking rules, we treat
slice-to-volume consolidation as a structured estimation problem
rather than a deterministic post-processing step. Instead of relying
on global averaging or extrema-based enclosure, our approach
integrates continuity constraints with robust coordinate statistics
and a lightweight axial refinement. Unlike 3D segmentation networks
that require dense voxel annotations unavailable in our clinical
setting, YOLO-PVC operates purely on bounding-box predictions,
providing a modular and annotation-efficient alternative for
volumetric localization in medical imaging.

\section{Method}

\subsection{Problem Formulation}

Let $V \in \mathbb{R}^{H \times W \times D}$ denote a volumetric image of spatial resolution $H \times W$ and depth $D$, represented as an ordered stack of 2D slices along the depth dimension. A 2D object detector $f_\theta$ is applied independently to each slice $z \in \{0, \dots, D-1\}$. For slice $z$, the detector produces a set of detections
\begin{equation}
B_z = \{(x_{1i}, y_{1i}, x_{2i}, y_{2i}, c_i)\}_{i=1}^{N_z},
\end{equation}
where $(x_{1i}, y_{1i})$ and $(x_{2i}, y_{2i})$ denote the top-left and bottom-right coordinates of the $i$-th bounding box, $c_i \in [0,1]$ denotes its confidence score, and $N_z$ is the number of detections in slice $z$. 

In practice, we retain detections belonging to the target class above 
a predefined confidence threshold $\tau$ and aggregate them across 
slices to produce a single volume-level prediction. The value of 
$\tau$ is selected empirically via ablation, as detailed in 
Section~\ref{sec:ablation}.

Our goal is to consolidate the slice-wise detections $\{B_z\}_{z=0}^{D-1}$ into a single axis-aligned 3D bounding box
\begin{equation}
B^{3D} = (x_{\min}, y_{\min}, z_{\min}, x_{\max}, y_{\max}, z_{\max}),
\end{equation}
that encloses the target object across the volume.

\begin{figure}[t]
    \centering
    \includegraphics[width=\linewidth]{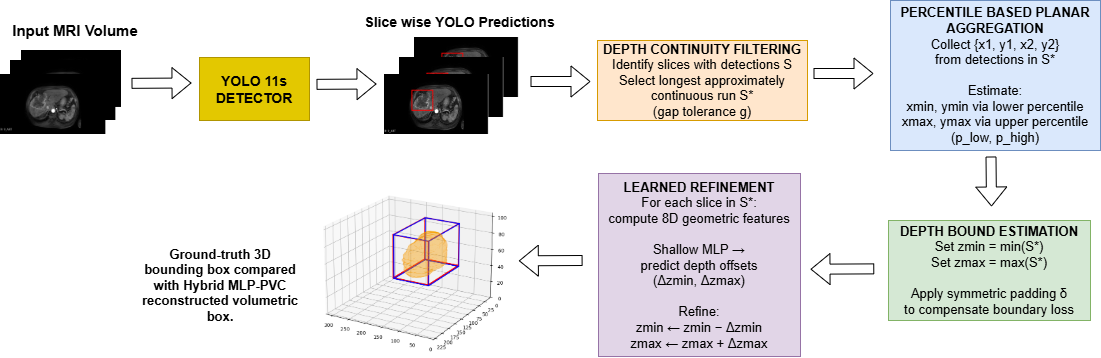}
    \caption{Overview of the proposed YOLO-PVC framework. Slice-wise detections are consolidated via depth continuity filtering, percentile-based planar aggregation, depth bound estimation, and MLP-based axial refinement to produce the final 3D bounding box.}
    \label{fig:pipeline}
\end{figure}

This consolidation problem is ill-posed due to slice-wise false positives, missed detections, geometric variability across depth, and unstable estimation of axial extent. Hence, we address these challenges using Percentile-Based Volumetric Consolidation (PVC), a lightweight procedure for aggregating slice-wise detections into a coherent 3D bounding box. An overview of the proposed consolidation pipeline is shown in Fig.~\ref{fig:pipeline}.

\subsection{Depth Continuity Filtering}

Let

\begin{equation}
S = \{ z \mid B_z \neq \emptyset \}
\end{equation}

denote the set of slice indices containing at least one detection. Because independent slice-wise inference may produce isolated false positives, we enforce depth continuity by selecting the longest approximately contiguous subset

\begin{equation}
S^* = \arg\max_{S_j \subset S} |S_j| 
\quad \text{s.t.} \quad z_{k+1} - z_k \le g,
\end{equation}

where $\{z_k\}$ denotes the sorted elements of $S_j$, and where $g \ge 0$ is a gap tolerance parameter controlling the maximum allowable discontinuity between consecutive slices. The value of $g$ is selected empirically via ablation, 
as detailed in Section~\ref{sec:ablation}.

This step suppresses isolated detections while tolerating limited missed slices caused by detector instability. The filtered index set $S^*$ defines the depth support over which geometric aggregation is performed.

\subsection{Percentile-Based Spatial Aggregation (PVC)}

Given detections within $S^*$, we aggregate planar coordinates using empirical percentiles. Let

{\small
\begin{equation}
\mathcal{X}_1 = \{x_{1i} \mid z \in S^*\}, \quad
\mathcal{Y}_1 = \{y_{1i} \mid z \in S^*\}, \quad
\mathcal{X}_2 = \{x_{2i} \mid z \in S^*\}, \quad
\mathcal{Y}_2 = \{y_{2i} \mid z \in S^*\}
\end{equation}
}

collect planar coordinates from all detections across slices in $S^*$.

We define the consolidated planar bounds as

\begin{equation}
x_{\min} = Q_{p_l}(\mathcal{X}_1), \quad
y_{\min} = Q_{p_l}(\mathcal{Y}_1),
\end{equation}

\begin{equation}
x_{\max} = Q_{p_u}(\mathcal{X}_2), \quad
y_{\max} = Q_{p_u}(\mathcal{Y}_2),
\end{equation}

where $Q_p(\cdot)$ denotes the empirical $p$-th percentile, and $0 < p_l < 0.5 < p_u < 1$.

Unlike Min--Max Stacking, which is sensitive to single extreme detections, or Median Fusion and Trimmed Mean Aggregation, which may bias extent estimation under asymmetric noise, percentile aggregation provides controlled robustness by retaining a predefined coverage of the spatial coordinate distribution. This formulation treats slice-wise detections as samples from a noisy geometric process and estimates bounds using robust order statistics rather than extrema. Specifically, the 10/90 percentile configuration
tolerates up to approximately 10\% coordinate contamination from false
positives or missed detections while preserving nearly complete tumor
coverage, as empirically confirmed by the ablation over
$p \in \{5/95, 10/90, 20/80\}$ in Section~\ref{sec:ablation}.

\subsection{Depth Padding}

Initial depth bounds are determined from the filtered slice indices:

\begin{equation}
z_{\min} = \min(S^*), \quad
z_{\max} = \max(S^*).
\end{equation}

Because boundary slices are often under-detected, we introduce symmetric padding:

\begin{equation}
z_{\min} = \max(0, z_{\min} - \delta), \quad
z_{\max} = \min(D-1, z_{\max} + \delta),
\end{equation}

where $\delta \ge 0$ is a padding parameter. The padding parameter $\delta$ is selected empirically 
via ablation, as detailed in Section~\ref{sec:ablation}.

Depth padding compensates for systematic underestimation near object boundaries and improves volumetric recall without modifying planar localization.

The resulting heuristic PVC estimate is therefore

\begin{equation}
B^{3D}_{PVC} = (x_{\min}, y_{\min}, z_{\min}, x_{\max}, y_{\max}, z_{\max}).
\end{equation}

\subsection{MLP-Based Calibration (MLP-PVC)}
\label{sec:mlp-pvc}

Although PVC provides robust statistical consolidation, residual depth bias may persist due to structured slice-dependent noise, particularly near object boundaries. To correct these systematic deviations, we introduce a lightweight calibration module operating on slice-level geometric descriptors.

For each slice $i \in S^*$, we compute an 8D feature vector
\begin{equation}
f_i \in \mathbb{R}^8,
\end{equation}
derived exclusively from predicted bounding-box geometry and confidence. The features encode normalized area, aspect ratio, confidence score, relative depth position within the selected run, inter-slice area variation, and normalized centroid deviation. As these features depend only on detector outputs, the calibration module remains modality-agnostic.

The calibration network is a shallow multi-head MLP with a shared trunk:
\begin{equation}
h_i = \phi(W_2\,\phi(W_1 f_i)),
\end{equation}
where $W_1:\mathbb{R}^{8}\!\rightarrow\!\mathbb{R}^{64}$ and $W_2:\mathbb{R}^{64}\!\rightarrow\!\mathbb{R}^{32}$ are fully connected layers with ReLU activations ($\phi$). Two linear heads (without activation) are applied to $h_i$:
\begin{equation}
w_i = W_w h_i \in \mathbb{R}, \qquad
\Delta z_i = W_z h_i \in \mathbb{R}^2,
\end{equation}
where $w_i$ represents a learned slice-importance score and $\Delta z_i = (\Delta z_{\min}^{(i)}, \Delta z_{\max}^{(i)})$ denotes per-slice depth offset predictions. In the current formulation, only the offset head is used for volumetric refinement. Each slice contributes through its predicted offset $\Delta z_i$, and slice contributions are aggregated uniformly via mean pooling.

Volume-level correction is obtained via mean aggregation over slices:
\begin{equation}
\bar{\Delta z} = \frac{1}{|S^*|}\sum_{i\in S^*}\Delta z_i.
\end{equation}

The two predicted offsets are averaged to obtain a symmetric depth expansion:
\begin{equation}
\Delta z_{\text{sym}} = \frac{1}{2}\left(\bar{\Delta z}_{\min} + \bar{\Delta z}_{\max}\right),
\end{equation}
which is clipped to a bounded range to prevent excessive expansion. The final depth bounds are refined as
\begin{equation}
z_{\min} \leftarrow z_{\min} - \Delta z_{\text{sym}}, \quad
z_{\max} \leftarrow z_{\max} + \Delta z_{\text{sym}}.
\end{equation}

Training supervision is derived from discrepancies between run boundaries and ground-truth depth limits:
\begin{equation}
t_{\min} = \max(0, z_{\min}^{\text{run}} - z_{\min}^{\text{GT}}), \quad
t_{\max} = \max(0, z_{\max}^{\text{GT}} - z_{\max}^{\text{run}}),
\end{equation}
with $t=(t_{\min}, t_{\max})$. Features are standardized using dataset statistics, and the model is optimized via Huber loss \cite{Huber1964}:
\begin{equation}
\mathcal{L} = \text{Huber}\!\left(\frac{1}{|S^*|}\sum_{i\in S^*}\Delta z_i,\; t\right).
\end{equation}

The MLP introduces negligible overhead ($\approx$2.7K parameters) and learns residual depth corrections while preserving the statistical and model-agnostic structure of PVC.

\section{Experimental Setup}

\subsection{Dataset}
\label{sec:dataset}

This retrospective study included patients with pathologically
confirmed primary liver malignancies who underwent surgical resection
at Henri Mondor University Hospital (Cr\'eteil, France). The cohort
comprised three tumor categories: hepatocellular carcinoma (HCC),
intrahepatic cholangiocarcinoma (CCA), and combined
hepatocellular--cholangiocarcinoma (cHCC--CCA), also referred to as
Mixed tumor. Preoperative contrast-enhanced MRI was
acquired on Siemens 3T (Skyra) and 1.5T (Avanto Fit) scanners.
Although the clinical protocol included multi-phase acquisitions, the
present study was restricted to the arterial phase, where tumors
typically exhibit strong hyperenhancement, improving lesion conspicuity and making this phase particularly suitable for
detection~\cite{Wang2023UNetPP}.

Among 147 identified patients, three-dimensional arterial-phase MRI was available for 142 volumes (85 HCC, 22 CCA, and 35 Mixed), while five cases were excluded due to missing arterial-phase acquisitions. Expert tumor delineations provided by radiologists were used to derive ground-truth 3D bounding boxes for evaluation and training label generation.

For YOLO training, arterial-phase volumes were converted into slice-wise 2D detection samples with corresponding bounding boxes. Dataset splitting was performed at the \emph{patient (case) level} to prevent data leakage, such that all slices from a given volume are assigned exclusively to a single split. Patients were stratified by tumor category and split into training/validation/test sets by independently shuffling cases within each class (fixed seed) and applying an 80/10/10 partition with integer rounding.

The final split contains 113 training cases (68 HCC, 17 CCA, 28 Mixed), 13 validation cases (8 HCC, 2 CCA, 3 Mixed), and 16 test cases (9 HCC, 3 CCA, 4 Mixed), totaling 142 volumes. This yields 2058 training slices, 211 validation slices, and 169 test slices (only slices with available labels are included). The slice-level class distribution is 1236 HCC, 307 CCA, and 515 Mixed slices in the training set; 103 HCC, 27 CCA, and 81 Mixed in validation; and 89 HCC, 32 CCA, and 48 Mixed in the test set. Each image has a corresponding YOLO-format label file with consistent class indexing (HCC: 0, CCA: 1, Mixed: 2).

\subsection{Slice-wise Detector Training}

We adopt a slice-wise 2D detection strategy to enable scalable volumetric modeling without training computationally intensive 3D networks. Specifically, we train Ultralytics YOLO11s initialized from publicly available pretrained weights. The YOLO11s variant offers a favorable balance between representational capacity and computational efficiency, while benefiting from stable optimization through pretrained initialization \cite{UltralyticsYOLO}.

Training is performed on the dataset described in Sec.~\ref{sec:dataset}. Input resolution is fixed to $512\times512$ with batch size 16 and 100 training epochs. Early stopping with patience 20 and fixed random seed 42 are used to ensure reproducibility. Mixed-precision training is enabled.

Optimization is handled by the Ultralytics training engine with automatic optimizer selection, resulting in AdamW with learning rate $1.429\times10^{-3}$ and weight decay $5\times10^{-4}$. Default Ultralytics data augmentation settings are used, including mosaic augmentation, horizontal flipping, HSV color perturbation, translation and scale jittering, random erasing, and lightweight Albumentations transforms (blur, median blur, grayscale conversion, and CLAHE). No additional custom augmentation strategies are introduced.

Experiments are implemented in PyTorch 2.9.0 with CUDA 12.6 and executed on an NVIDIA L4 GPU (22.7\,GB).

\subsection{MLP-PVC Training}
 To refine depth extent estimation beyond fixed heuristics, we train the calibration module described in \cref{sec:mlp-pvc} using volumes from the training split only. For each volume, slice-wise YOLO predictions are filtered by class and the longest contiguous detection run is extracted (gap tolerance $g=1$). For each run of length $N_i$, slice-level 8D geometric features are computed, yielding a feature matrix of size $[N_i, 8]$.

Supervision is defined as the depth gaps between run boundaries and 
the ground-truth depth extent: $t_{\min}=\max(0,z_{\min}^{\text{run}}-z_{\min}^{\text{GT}})$ and $t_{\max}=\max(0, z_{\max}^{\text{GT}}-z_{\max}^{\text{run}})$. Features are standardized using dataset-wide mean and standard deviation computed over all training runs. The network is trained for 400 epochs using Adam with learning rate 0.005 and weight decay $10^{-5}$. We optimize a Huber loss between the mean-pooled predicted gaps and the target gaps, with gradient clipping (max norm 5) for stability. The calibration module contains approximately 2.7K parameters and introduces negligible overhead relative to slice-wise YOLO inference.

\subsection{Evaluation Metrics}

Predicted axis-aligned 3D bounding boxes are evaluated using complementary overlap and localization metrics. IoU$_{3D}$ is our primary metric, as it directly measures volumetric agreement, or volumetric intersection-over-union, between predicted and ground-truth boxes. It is particularly unforgiving: small misalignments along any axis, especially depth, substantially reduce intersection volume \cite{Ming2023DeepDI}, making IoU$_{3D}$ a strict and reliable indicator of spatial consistency.

The volumetric Dice coefficient (Dice$_{3D}$) is also reported as a smoother overlap measure widely used in medical imaging, providing complementary insight \cite{Kern2020}. Unlike IoU, Dice places slightly less penalty on small boundary mismatches and is therefore useful for interpreting overlap consistency when volumetric agreement is high but minor extent deviations exist.

To isolate planar localization from depth errors, we compute Bird’s-Eye-View IoU (BEV IoU) by projecting both 3D boxes onto the $xy$ plane and measuring 2D overlap \cite{Yan2018SECOND}. This separates lateral localization quality from axial consistency.

Centroid distance is measured as the Euclidean distance between predicted and ground-truth box centers in voxel space to quantify global localization accuracy \cite{Ebner2020FetalBrainSRR}.

Finally, per-axis dimension errors $(\Delta x, \Delta y, \Delta z)$ quantify systematic over- or underestimation of box extent. In particular, $\Delta z$ captures axial bias introduced by slice-wise aggregation \cite{Geiger2012}.

Together, these metrics provide a concise yet comprehensive evaluation of volumetric overlap, localization accuracy, and depth consistency.

\subsection{Computational Efficiency}

All measurements are performed on an NVIDIA L4 GPU (CUDA) in inference mode. YOLO11s (9.41M parameters) requires 10.00 ms per $512\times512$ slice on average (p50: 9.90 ms, p90: 10.20 ms, p95: 10.45 ms), with peak GPU memory usage of 95.36 MB. For a representative volume of $N=80$ slices, end-to-end slice-wise inference takes 799.6 ms on average (p95: 820.3 ms). PVC introduces no learnable parameters, and the MLP calibration module adds approximately 2.7K parameters ($\approx 0.03\%$ overhead) with negligible additional runtime.

\section{Results and Discussion}

\subsection{Comparison with 2D-to-3D Aggregation Baselines}

Table~\ref{tab:baseline_overall} reports overall volumetric performance under the final evaluation protocol. 
We evaluate 3D IoU, 3D Dice, BEV IoU (axial-plane overlap of the projected 3D box), centroid distance (in voxels), and axis-wise size errors ($\Delta x, \Delta y, \Delta z$). All metrics are averaged across tumor classes. Slice-wise detection followed by cross-slice consolidation is a common paradigm for deriving 3D bounding boxes from 2D predictions in volumetric medical imaging pipelines. The five baseline aggregation strategies are defined in Section~\ref{sec:relatedwork}.

\begin{table}[t]
\centering
\caption{Overall volumetric performance under patient-level split (final evaluation).}
\label{tab:baseline_overall}
\resizebox{\linewidth}{!}{
\begin{tabular}{lccccccc}
\toprule
Method & IoU$_{3D}$ & Dice$_{3D}$ & BEV IoU & Centroid & $\Delta x$ & $\Delta y$ & $\Delta z$ \\
\midrule
Confidence-Weighted Avg.        & 0.424 & 0.562 & 0.591 & 13.878 & -9.272 & -8.926 & -4.507 \\
Trimmed Mean Aggregation        & 0.429 & 0.566 & 0.597 & 14.048 & -8.779 & -8.574 & -4.507 \\
Median Fusion                   & 0.457 & 0.590 & 0.637 & 13.380 & -7.397 & -7.250 & -4.507 \\
Continuity-Aware Linking        & 0.583 & 0.682 & 0.782 & 12.571 & +1.022 & +0.588 & -4.507 \\
Min--Max Stacking               & 0.596 & 0.689 & 0.795 & 13.238 & +6.471 & +2.787 & -4.507 \\
\midrule
PVC (Heuristic)                 & 0.665 & 0.745 & 0.781 & 12.674 & +1.243 & +0.647 & -2.610 \\
Hybrid MLP-PVC                  & 0.710 & 0.781 & 0.782 & 12.587 & +1.022 & +0.588 & -0.125 \\
\bottomrule
\end{tabular}
}
\end{table}

The three purely coordinate-fusion baselines (confidence-weighted, trimmed mean, and median fusion) substantially underperform (IoU$_{3D}$=0.424--0.457). Although statistically robust, they do not enforce inter-slice continuity, leading to accumulated lateral drift, unstable depth extent, and systematic axial underestimation ($\Delta z=-4.51$), with large centroid errors ($\approx$13--14 voxels).

Introducing explicit cross-slice association via continuity-aware linking markedly improves volumetric overlap (IoU$_{3D}$=0.583), confirming that geometric continuity is essential for stable 3D reconstruction. Min--Max stacking further increases IoU$_{3D}$ to 0.596 and achieves the strongest BEV IoU (0.795) through its enclosure strategy. However, both linking and Min--Max stacking retain severe axial bias ($\Delta z=-4.51$), demonstrating that continuity enforcement alone does not correct depth underestimation.

The proposed heuristic PVC replaces global coordinate fusion with structured consolidation. By selecting the longest contiguous detection run, applying percentile-based lateral aggregation, and introducing controlled axial padding, PVC improves IoU$_{3D}$ from 0.596 (Min--Max stacking) to 0.665, corresponding to a \textbf{+11.6\% relative improvement}, and increases Dice from 0.689 to 0.745. Depth bias is reduced from $\Delta z=-4.51$ to $-2.61$, representing a \textbf{42\% reduction}, confirming that robust geometric estimation across slices substantially stabilizes volumetric reconstruction.

Hybrid MLP-PVC further replaces fixed axial padding with a lightweight learned calibration module modeling residual depth discrepancies. Without altering lateral fusion, this refinement increases IoU$_{3D}$ to 0.710, corresponding to a \textbf{+19.2\% relative improvement over Min--Max stacking} and a \textbf{+6.8\% improvement over heuristic PVC}, while shifting axial bias toward near-zero ($\Delta z=-0.125$). These results demonstrate that structured geometric consolidation combined with minimal learned axial correction substantially improves volumetric reconstruction within slice-wise detection pipelines while remaining computationally efficient.

\subsection{Per-Class Analysis}

To evaluate class-specific behavior under heterogeneous tumor morphology, we report per-class performance for the two strongest configurations: heuristic PVC and Hybrid MLP-PVC (Table~\ref{tab:perclass_final}). We restrict per-class reporting to these two variants because they represent the best-performing heuristic and learned formulations, respectively, and clearly outperform all classical aggregation baselines in overall evaluation (Table~\ref{tab:baseline_overall}). This focused comparison isolates the effect of learned depth calibration across clinically relevant tumor types.

\begin{table}[t]
\centering
\caption{Per-class volumetric performance under patient-level split.}
\label{tab:perclass_final}
\scriptsize
\setlength{\tabcolsep}{4pt}
\resizebox{\linewidth}{!}{
\begin{tabular}{l l c c c c c c c}
\toprule
Class & Method & IoU$_{3D}$ & Dice & BEV IoU & Centroid & $\Delta x$ & $\Delta y$ & $\Delta z$ \\
\midrule
HCC   & PVC    & 0.654 & 0.730 & 0.759 & 16.42 & +3.70 & +3.50 & -2.10 \\
HCC   & Hybrid & 0.697 & 0.763 & 0.759 & 16.30 & +3.37 & +3.42 & +0.35 \\
\midrule
CCA   & PVC    & 0.733 & 0.834 & 0.870 & 9.85 & -3.20 & -4.00 & -1.80 \\
CCA   & Hybrid & 0.804 & 0.889 & 0.869 & 9.80  & -3.21 & -4.05 & +0.37 \\
\midrule
Mixed & PVC    & 0.653 & 0.731 & 0.788 & 11.74 & -2.50 & -3.90 & -4.20 \\
Mixed & Hybrid & 0.687 & 0.764 & 0.788 & 11.66 & -2.52 & -3.94 & -1.61 \\
\bottomrule
\end{tabular}
}
\end{table}

Across all tumor types, Hybrid MLP-PVC consistently improves volumetric overlap. For HCC, IoU$_{3D}$ increases from 0.654 to 0.697 (+6.6\% relative), and Dice from 0.730 to 0.763. For CCA, IoU$_{3D}$ improves from 0.733 to 0.804 (+9.7\% relative), and for Mixed tumors from 0.653 to 0.687 (+5.2\% relative). These gains are systematic across morphology categories, indicating that the learned axial calibration generalizes beyond a single tumor phenotype.

The most notable improvement occurs along the depth axis. Under heuristic PVC, all classes exhibit residual negative $\Delta z$ values, reflecting incomplete superior--inferior coverage (e.g., $-2.1$ for HCC and $-4.2$ for Mixed). Hybrid MLP-PVC shifts $\Delta z$ toward near-unbiased reconstruction (+0.35 for HCC, +0.37 for CCA), and substantially reduces axial underestimation for Mixed tumors ($-1.61$). Importantly, this depth correction does not degrade lateral stability: $\Delta x$ and $\Delta y$ remain comparable, and BEV IoU remains stable (0.759--0.870 range).

Per-class analysis is particularly important in abdominal MRI, where tumor types differ in shape regularity, boundary sharpness, and contrast behavior. The consistent improvements across HCC, CCA, and Mixed tumors indicate that percentile-based consolidation captures robust lateral structure, while the lightweight MLP calibration adapts axial extent according to slice-level reliability patterns rather than fixed padding rules. This supports the hypothesis that minimal learned depth reasoning enhances volumetric reconstruction in heterogeneous clinical scenarios.

\begin{figure*}[t]
    \centering
    \includegraphics[width=\textwidth]{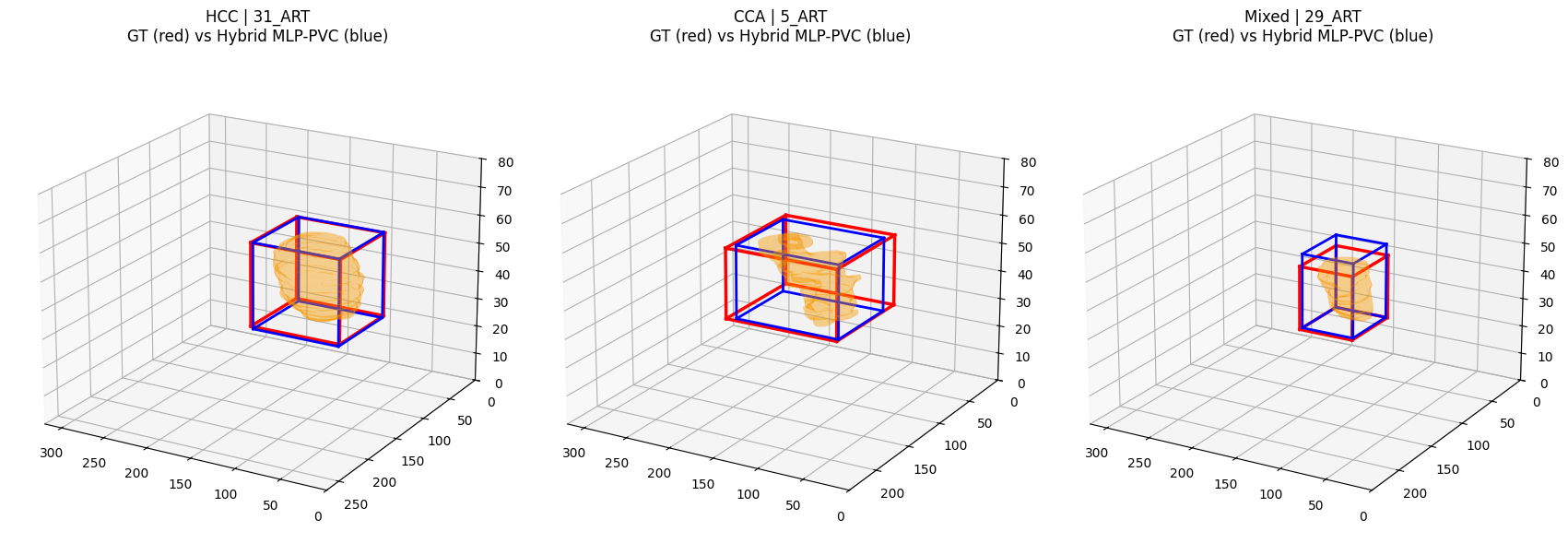}
    \caption{Qualitative comparison of volumetric localization on three randomly selected cases, one from each tumor category (HCC, CCA, and Mixed). Ground-truth 3D bounding boxes are shown in red, and Hybrid MLP-PVC reconstructed boxes are shown in blue. Across varying tumor morphologies and spatial extents, the reconstructed volumes exhibit close geometric alignment with ground truth, particularly along the depth axis, illustrating stable slice-to-volume consolidation under the proposed framework.}
    \label{fig:qualitative_results}
\end{figure*}

\subsection{Ablation Studies and Robustness Analysis}
\label{sec:ablation}

Table~\ref{tab:ablation_final} summarizes the comprehensive \emph{Ablation Studies} over all core design components: inter-slice gap tolerance ($g$), percentile bounds ($p$), confidence threshold ($\tau$), depth padding ($\delta$), and hybrid MLP refinement. All results are reported under the same evaluation protocol for direct comparability.

Enforcing inter-slice continuity is critical for volumetric stability. When gap tolerance is disabled ($g=0$), performance degrades severely (IoU$_{3D}$=0.036), confirming that naive stacking without continuity constraints is highly unstable in volumetric reconstruction. Moderate tolerances ($g=1$) restore stable behavior (IoU$_{3D}$=0.668), while larger tolerances ($g=2,3$) slightly degrade accuracy due to inclusion of noisy slices. This highlights the importance of structured slice connectivity.

Percentile-based lateral fusion exhibits a clear robustness trade-off. Conservative bounds (20/80) reduce overlap (IoU$_{3D}$=0.644) due to excessive trimming of valid boundary slices, whereas more aggressive bounds (5/95) slightly improve IoU (0.675) at the cost of increased lateral drift. The default 10/90 configuration provides a stable balance between robustness and overlap consistency.

Confidence thresholding reveals a stability--recall trade-off. Lower thresholds ($\tau=0.25$) increase centroid error and reduce IoU (0.648), indicating that low-confidence detections destabilize consolidation. Higher thresholds ($\tau=0.70$) improve centroid precision but slightly reduce volumetric consistency. The intermediate setting ($\tau=0.50$) achieves balanced performance across metrics.

Depth padding has the strongest impact on volumetric accuracy. Removing padding ($\delta=0$) leads to pronounced axial underestimation (IoU$_{3D}$=0.583, $\Delta z=-4.507$), confirming systematic boundary slice omission. Moderate padding ($\delta=1$) restores stable performance (IoU$_{3D}$=0.668), while $\delta=2$ further improves IoU$_{3D}$ to 0.705 and reduces depth bias ($\Delta z=-0.544$). This indicates that axial coverage strongly influences 3D box reconstruction quality.

Finally, replacing heuristic consolidation with Hybrid MLP-PVC yields the strongest performance (IoU$_{3D}$=0.710) while shifting residual depth bias toward near-zero ($\Delta z=-0.125$). Importantly, these gains are achieved without degrading BEV IoU or lateral stability, indicating that learned refinement enhances axial calibration rather than distorting planar geometry.

Overall, the ablation confirms that continuity enforcement, percentile fusion, confidence gating, depth padding, and learned refinement each contribute measurably to volumetric reconstruction quality. The selected configuration emerges as a principled balance between robustness, overlap accuracy, and depth stability, supporting the structural design of the proposed PVC framework.

\begin{table*}[t]
\centering
\caption{Comprehensive ablation study under the final evaluation setup. Results are reported as overall mean across all volumes.}
\label{tab:ablation_final}
\resizebox{\textwidth}{!}{
\begin{tabular}{lccccccc}
\toprule
Setting & IoU$_{3D}$ & Dice$_{3D}$ & BEV IoU & Centroid & $\Delta x$ & $\Delta y$ & $\Delta z$ \\
\midrule
A\_g=0      & 0.036 & 0.063 & 0.261 & 50.374 & -16.926 & -15.699 & -18.574 \\
A\_g=1      & 0.668 & 0.747 & 0.782 & 12.567 &  +1.022 &  +0.588 &  -2.515 \\
A\_g=2      & 0.664 & 0.744 & 0.781 & 11.805 &  +2.309 &  +1.625 &  -2.154 \\
A\_g=3      & 0.656 & 0.739 & 0.773 & 11.919 &  +2.743 &  +2.118 &  -1.904 \\
\midrule
B\_p=20/80  & 0.644 & 0.733 & 0.759 & 12.493 &  -1.162 &  -1.471 &  -2.515 \\
B\_p=10/90  & 0.668 & 0.747 & 0.782 & 12.567 &  +1.022 &  +0.588 &  -2.515 \\
B\_p=5/95   & 0.675 & 0.751 & 0.789 & 12.485 &  +2.728 &  +1.625 &  -2.515 \\
\midrule
C\_$\tau$=0.25 & 0.648 & 0.725 & 0.752 & 13.707 &  +6.297 &  +2.391 &  -1.891 \\
C\_$\tau$=0.50 & 0.668 & 0.747 & 0.782 & 12.567 &  +1.022 &  +0.588 &  -2.515 \\
C\_$\tau$=0.70 & 0.676 & 0.765 & 0.812 & 10.078 &  -1.197 &  -1.439 &  -3.091 \\
\midrule
D\_$\delta$=0 & 0.583 & 0.682 & 0.782 & 12.571 &  +1.022 &  +0.588 &  -4.507 \\
D\_$\delta$=1 & 0.668 & 0.747 & 0.782 & 12.567 &  +1.022 &  +0.588 &  -2.515 \\
D\_$\delta$=2 & 0.705 & 0.773 & 0.782 & 12.565 &  +1.022 &  +0.588 &  -0.544 \\
\midrule
E\_HeuristicPVC      & 0.665 & 0.745 & 0.781 & 12.674 &  +1.243 &  +0.647 &  -2.610 \\
E\_Hybrid\_MLP\_PVC  & 0.710 & 0.781 & 0.782 & 12.587 &  +1.022 &  +0.588 &  -0.125 \\
\bottomrule
\end{tabular}
}
\end{table*}

For \emph{Robustness Analysis}, we evaluate the final Hybrid MLP-PVC under two stress conditions: (i) random removal of 10\% of slice detections and (ii) injection of three adversarial outlier slices per volume. Table~\ref{tab:robustness_final} reports the overall results. Under the standard evaluation protocol, the baseline achieves strong volumetric localization (IoU$_{3D}$=0.710, Dice=0.781) with stable centroid localization (12.587 voxels) and near-unbiased depth estimation ($\Delta z=-0.125$), consistently across HCC, CCA, and Mixed tumors.

\begin{table}[t]
\centering
\caption{Robustness evaluation under degraded detection conditions (patient-level split).}
\label{tab:robustness_final}
\resizebox{\linewidth}{!}{
\begin{tabular}{lccccccc}
\toprule
Condition & IoU$_{3D}$ & Dice$_{3D}$ & BEV IoU & Centroid & $\Delta x$ & $\Delta y$ & $\Delta z$ \\
\midrule
Baseline (Hybrid) & 0.710 & 0.781 & 0.782 & 12.587 & +1.022 & +0.588 & -0.125 \\
Missing 10\%      & 0.663 & 0.750 & 0.780 & 12.135 & +0.662 & -0.088 & -2.765 \\
Outliers +3       & 0.390 & 0.461 & 0.420 & 37.727 & +77.971 & +42.809 & -2.515 \\
\bottomrule
\end{tabular}
}
\end{table}

Under random slice removal, performance degrades moderately yet remains stable. IoU decreases from 0.710 to 0.663 ($-$4.7 percentage points), while BEV IoU is largely preserved (0.782 → 0.780) and centroid error changes only marginally (12.587 → 12.135). These bounded variations indicate that continuity enforcement and percentile-based fusion tolerate moderate axial sparsity and preserve geometric coherence across volumes.

In contrast, the second condition constitutes an intentional
adversarial stress test: three structured false-positive detections
are simultaneously injected per volume, a scenario significantly more
severe than typical clinical noise. The observed IoU drop from 0.710
to 0.390 is driven by extreme lateral expansion ($\Delta x$,
$\Delta y$) from outlier boxes, not by failure of the continuity or
percentile mechanisms themselves. Sensitivity to systematic structured
outliers is inherent to any geometric aggregation method. Under
realistic missing-detection conditions, the framework degrades
gracefully while maintaining structural stability.

\section{Conclusion}
This work demonstrates that slice-to-volume reconstruction is
primarily constrained by axial instability rather than planar
localization error. By enforcing inter-slice continuity and
percentile-based geometric aggregation, PVC reframes consolidation
as a structured estimation problem, yielding consistent improvements
in IoU$_{3D}$ and reduced depth bias. The hybrid MLP refinement
enhances axial calibration through lightweight learned correction,
achieving strong volumetric overlap without memory-intensive 3D
architectures. The framework operates on slice-level geometric
detections, independent of image intensities or modality-specific
features, supporting clinical deployment under annotation constraints.
Consistent improvements across tumor types confirm clinical relevance
for heterogeneous liver malignancies. Future work includes
multi-center validation and a fully integrated 3D detection pipeline.

%
%
\bibliographystyle{splncs04}
\bibliography{main}
\end{document}